\documentclass[letterpaper, 10pt]{article}

\usepackage{graphicx}
\usepackage{float}
\usepackage[inner=0.75in, outer=0.75in, top=0.75in, bottom=1in, paperheight=11in, paperwidth=8.5in]{geometry}
\usepackage{array}
\usepackage{rotating}
\usepackage{multirow}
\usepackage{mathtools}
\usepackage{shuffle}
\usepackage[table, svgnames, dvipsnames,xcdraw]{xcolor}
\usepackage[square, numbers, comma, sort&compress]{natbib}
\usepackage{verbatim}
\usepackage{bm}
\usepackage{amsmath}
\usepackage{amsfonts}
\usepackage{amssymb}
\usepackage{amsthm}
\usepackage{epstopdf}
\usepackage[utf8]{inputenc}
\usepackage[english]{babel}
\usepackage{color}
\usepackage{forest}
\usepackage{xcolor}
\usepackage{pdfpages}
\usepackage{pgfplots}
\usepackage{pgfplotstable}
\usepgfplotslibrary{patchplots}
\pgfplotsset{compat=1.15}
\usepackage{microtype}
\usepackage{url}
\usepackage{subfiles}
\usepackage{caption}
\usepackage{subcaption}
\usepackage{parskip}
\usepackage[section]{placeins}
\usepackage{tikz}
\usepackage{tikz-cd}
\usetikzlibrary{arrows,calc,shapes,decorations.pathreplacing}
\usetikzlibrary{decorations.markings}
\usetikzlibrary{trees,positioning,fit,shapes}
\usetikzlibrary{backgrounds}
\usetikzlibrary{patterns, positioning, arrows}
\usepackage{framed}
\colorlet{shadecolor}{blue!20}
\usepackage[obeyFinal,textwidth=0.55in]{todonotes}
\usepackage[hidelinks]{hyperref}
\usepackage[capitalize]{cleveref}
\usepackage{enumitem}

\theoremstyle{plain}
\newtheorem{thm}{Theorem}[section]
\theoremstyle{definition}
\newtheorem{defn}[thm]{Definition}

\makeatletter
\newcommand\setveclength[3]{
  \pgfpointdiff{\pgfpointanchor{#2}{center}}{\pgfpointanchor{#3}{center}}
  \pgfmathveclen{\pgf@x}{\pgf@y}
  \edef#1{\pgfmathresult}
}
\makeatother

\makeatletter
\renewenvironment{abstract}{%
    \if@twocolumn
      \section*{\abstractname}%
    \else 
      \begin{center}%
        {\bfseries \Large\abstractname\vspace{\z@}}
      \end{center}%
      \quotation
    \fi}
    {\if@twocolumn\else\endquotation\fi}
\makeatother

\usepackage{times}

\title{\bf On topological data analysis for structural dynamics: an introduction to persistent homology}
\author{{\bf T.\ Gowdridge, N.\ Dervilis, K.\ Worden} \\
        Dynamics Research Group, Department of Mechanical Engineering, University of Sheffield        \\
        Mappin Street,      \\
        Sheffield, S1 3JD,  \\
        UK                  \\
        tgowdridge1@sheffield.ac.uk
}

\begin{document}	
\maketitle

\begin{abstract}
Topological methods can provide a way of proposing new metrics and methods of scrutinising data, that otherwise may be overlooked. In this work, a method of quantifying the shape of data, via a topic called \textit{topological data analysis} will be introduced. The main tool within \textit{topological data analysis} (TDA) is \textit{persistent homology}. Persistent homology is a method of quantifying the shape of data over a range of length scales. The required background and a method of computing persistent homology is briefly discussed in this work.

Ideas from topological data analysis are then used for nonlinear dynamics to analyse some common attractors, by calculating their embedding dimension, and then to assess their general topologies. A method will also be proposed, that uses topological data analysis to determine the optimal delay for a time-delay embedding. TDA will also be applied to a Z24 Bridge case study in structural health monitoring, where it will be used to scrutinise different data partitions, classified by the conditions at which the data were collected. A metric, from topological data analysis, is used to compare data between the partitions. The results presented demonstrate that the presence of damage alters the manifold shape more significantly than the effects present from temperature.
\end{abstract}

\section{Introduction}
Topological methods are very rarely used in structural dynamics generally, although considering the structure and topology of observed data may be useful. Topological methods can provide new metrics and methods of scrutinising data; the most rudimentary and powerful of which, \textit{persistence homology}, will be discussed and used extensively in this paper. By applying topological methods, an understanding of the topological structure of data can be used to formulate arguments and develop an understanding of system parameters on a manifold shape. This statement is especially true when working with higher-dimensional data sets, where an intuitive understanding of a manifold is not easy to visualise. By using topological methods, an understanding can be quantifiably analysed. When considering engineering data, data-sets are often embedded in higher-dimensional space, and topological information is often not well explored, leaving out potentially important and insightful information.

Topological data analysis (TDA), is a recently-developed and fast-growing field that has found its way into many areas of science and engineering. The general idea of TDA applies concepts from algebraic topology to data sets. The primary focus of TDA is to determine the shape of the manifold in which sampled data are embedded. This process is achieved by identifying $2D$ holes, $3D$ cavities and higher-dimensional analogues within the data structures. From these sampled data, an approximation to the topological structure can be calculated by use of \textit{simplicial complexes}, which are higher-dimensional analogues of graphs. From the simplicial complexes, the persistent homology can be calculated, this is then used to understand the topological structure of the data. The persistent homology is invariant for each data set, and can be used to identify the data set; much like a fingerprint. 

After the simplicial complexes have been constructed for some point data, these can be manipulated using ideas from algebraic topology, in order to determine algebraic groups that will capture information about the shape and structure of the simplicial complex; these groups are topological invariants. For any simplicial complex, algebraic topology can deduce a property called the \textit{homology}, which encodes information  about its number of $k-$dimensional holes. A generalisation of homology will be discussed here with respect to pioneering work by Edelsbrunner \cite{edelsbrunner2010computational, edelsbrunner2000topological}, where the homology can be considered over a range of simplicial complexes; this is called the \textit{persistent homology}, aptly named, as this uncovers how the homology persists over a range of scales.

The field of algebraic topology is well studied, but the application of the ideas to discrete point clouds has contributed to a boom in computational topology. Many standard packages exist to analyse data with regards to topological methods: GUDHI \cite{maria2014gudhi} and Ripser \cite{Bauer2021Ripser} being the most influential throughout this piece of work.

The layout of the paper is as follows: Section 2 will be devoted to persistent homology and its significance, and will provide an intuitive understanding. Some use cases of persistent homology will also be discussed. Section 3 will introduce and analyse the topology of some common attractors from the field of nonlinear dynamics. Section 4 will then go on to look at the classic Z24 Bridge structural health monitoring (SHM) case study, and explore the role of topology as a metric between partitions of the data set.

\section{Topological Data Analysis}
Only the strict mathematical formulations will be introduced here. For more information regarding the mathematics of group theory, and algebraic topology, the authors would advise referring to \cite{intrototopologybertmendelsen, nash1988topology, ghrist2018homological, maclane2012homology}, and \cite{genomics, ghrist2014EAT, boissonnat2018geometric, chazal2017introduction, edelsbrunner2000topological, zomorodian2005topology, carlsson2019persistent, edelsbrunner2010computational, ghrist2008barcodes} for more on TDA. For some other interesting applications in economics and genomics, the reader can refer to \cite{economicsTDA, diabetes, genomics}.

\subsection{Manifolds}
Manifolds are continuous surfaces from which the data are assumed to be sampled. By understanding the topology of the sampled data points, it is the aim of TDA to extract topological information about the underlying manifold from the data. The manifold shape is unknown prior to analysis, and persistent homology will identify features within the manifold over a range of length scales. Thereby, understanding the shape of the sampled data, it is the conjecture of TDA that the shape of the manifold is also understood.

Formally, a manifold is a space that is locally homeomorphic to some $n-$dimensional Euclidean space, $\mathbb{R}^n$. In this work, manifolds are primarily thought of as the underlying space from which the generated or collected data are sampled. Throughout this work, there will be a reoccurring idea that a change in the parameters of a dynamic system may change the shape of the associated manifold, therefore the topology can potentially be used to identify a change in the system.

\subsection{Simplicial Complex}
Simplicial complexes will be used as a way of attributing quantifiable shape to the data; they can be thought of as higher-dimensional analogues of graphs, giving a way of encoding connections between vertices. In TDA, the vertices of the simplicial complexes are the observed data points. Simplicial complexes can be analysed to output the homology of the data, and following this, the persistent homology. The homology and persistent homology are key topological invariants that can be used to describe the structure of the data.

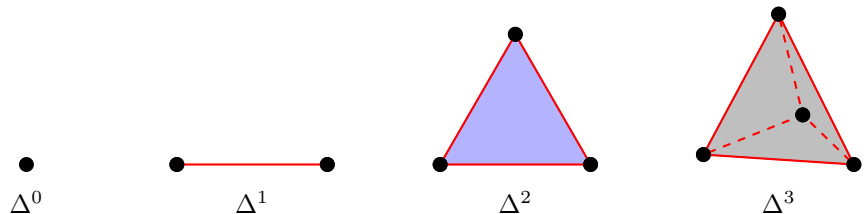
\begin{figure}[H]
    \centering
    
\begin{tikzpicture}[ele/.style={fill=black,circle,minimum width=3pt,inner sep=2pt}]
\node[ele] (a1) at (0,0) {};
\node at (0,-0.5) {$\Delta^0$};

\begin{scope}[shift={(2,0)}]
    \node[ele] (a1) at (0,0) {};    
    \node[ele] (a2) at (2,0) {};
    \draw[-,thick,shorten <=2pt,shorten >=2pt, red] (a1.center) -- (a2.center);
    \node at (1,-0.5) {$\Delta^1$};
    
    \node[ele] (a1) at (0,0) {};    
    \node[ele] (a2) at (2,0) {};
\end{scope}

\begin{scope}[shift={(5.5,0)}]
    \node[ele] (a1) at (2,0) {};    
    \node[ele] (a2) at (0,0) {};
    \node[ele] (a3) at (1,1.732) {};

    \draw[-,thick,shorten <=2pt,shorten >=2pt, red] (a1.center) -- (a2.center);
    \draw[-,thick,shorten <=2pt,shorten >=2pt, red] (a1.center) -- (a3.center);
    \draw[-,thick,shorten <=2pt,shorten >=2pt, red] (a2.center) -- (a3.center);

    \begin{scope}[on background layer]
        \path [fill=blue!30, draw] (a1.center) to (a2.center) to (a3.center) to (a1.center);
    \end{scope} 

    \node at (1,-0.5) {$\Delta^2$};
    \node[ele] (a1) at (2,0) {};    
    \node[ele] (a2) at (0,0) {};
    \node[ele] (a3) at (1,1.732) {};
    
\end{scope}

\begin{scope}[shift={(9,0)}]
    \node[ele] (a1) at (2,0) {};    
    \node[ele] (a2) at (0,0.132) {};
    \node[ele] (a3) at (1.32,0.66) {};
    \node[ele] (a4) at (1,2) {};
    
    \draw[-,thick,shorten <=2pt,shorten >=2pt, red] (a1.center) -- (a2.center);
    \draw[-,thick,shorten <=2pt,shorten >=2pt, red] (a1.center) -- (a4.center);
    \draw[-,thick,shorten <=2pt,shorten >=2pt, red] (a2.center) -- (a4.center);

    \draw[-,thick,shorten <=2pt,shorten >=2pt, dashed, red] (a1.center) -- (a3.center);
    \draw[-,thick,shorten <=2pt,shorten >=2pt, dashed, red] (a2.center) -- (a3.center);
    \draw[-,thick,shorten <=2pt,shorten >=2pt, dashed, red] (a4.center) -- (a3.center);
    
    \begin{scope}[on background layer]
        \path [fill=lightgray, draw] (a1.center) to (a2.center) to (a4.center) to (a1.center);
    \end{scope} 
\node at (1,-0.5) {$\Delta^3$};
    \node[ele] (a1) at (2,0) {};    
    \node[ele] (a2) at (0,0.132) {};
    \node[ele] (a3) at (1.32,0.66) {};
    \node[ele] (a4) at (1,2) {};

\end{scope}
\end{tikzpicture}
    \caption{The first four simplices.}
    \label{fig:simplices}
\end{figure}

A simplicial complex is a structure made up of fundamental building blocks called \textit{simplices}, the first four types of which are shown in Fig. \ref{fig:simplices}. Each vertex in the simplex is fully connected to all the other vertices and the space enclosed by the vertices is part of that simplex. For instance, $\Delta^2$ encloses a two-dimensional area, $\Delta^3$ encloses a three-dimensional volume. This sequence can be generalised for $\Delta^k$ enclosing a $k-$dimensional space between $(k+1)$ fully connected vertices.

A \textit{filtration} of a simplicial complex, $K$, is a nested sequence of sub-complexes, such that, $\emptyset = K^0 \subset K^1 \subset ... \subset K^m = K$, where $K^{i+1} = K^i \cup \Delta^i$ where $\Delta^i$ is a simplex of $K$ \cite{boissonnat2018geometric}, and $\emptyset$ is the empty set, i.e, there are no points inside.

There are many ways to construct a simplicial complex from point data. For simplification, only one method will be discussed within this paper, the \textit{Vietoris-Rips} (VR) complex \cite{carlsson2006algebraic}. The VR complex can be constructed for point data present in a \textit{metric space}.

A \textit{metric} $\partial_X$ is defined on a set $X$, which maps two elements from $X$ into the positive real numbers. The mapping gives the associated distance between the two elements. For a set to be deemed a metric space, it must satisfy the following axioms:
\begin{enumerate}
    \item for any $x, y \in X$ then $\partial_X(x, y) = \partial_X(y, x)$;
    \item for any $x, y \in X$ then $\partial_X(x,y) \geq 0 $, and $=0$ iff $x = y$;
    \item for any $x, y, z \in X$ then $\partial_X(x,z) \leq \partial_X(x,y) + \partial_X(y,z)$. 
\end{enumerate}
If these axioms are satisfied, $(X, \partial_X)$ forms a metric space, where $\partial_X$ is the metric on $X$.

An \textit{open ball} is defined on a metric space, $(X, \partial_X)$. For a point $a \in X$ and $\varepsilon > 0$, the subset of $X$ consisting of all the points $x \in X$ such that $\partial_X(a,x) < \varepsilon$ is referred to as the open ball of radius $\varepsilon$ at $a$.

For the VR complex $\text{VR}_\varepsilon(X,\partial_X)$, let $(X,\partial_X)$ be a finite metric space and $\varepsilon \in \mathbb{R}_{>0}$ be a fixed value that determines the scale of the VR complex \cite{chambers2010vietoris}, where $X$ is the set of vertices and $\partial_X$ is the metric on $X$. A VR complex is defined by the following condition

\begin{equation}
    \{x_0, ..., x_k\} \in \text{VR}_\varepsilon(X, \partial_X) \Leftrightarrow \|x_i - x_j\| \leq \varepsilon , \ \forall i,j \in \{0,...,k\}.
\end{equation}

As $\varepsilon$ goes from $0$ to $+\infty$, a nested sequence of complexes defines the \textit{Vietoris-Rips filtration} \cite{boissonnat2018geometric}.
 
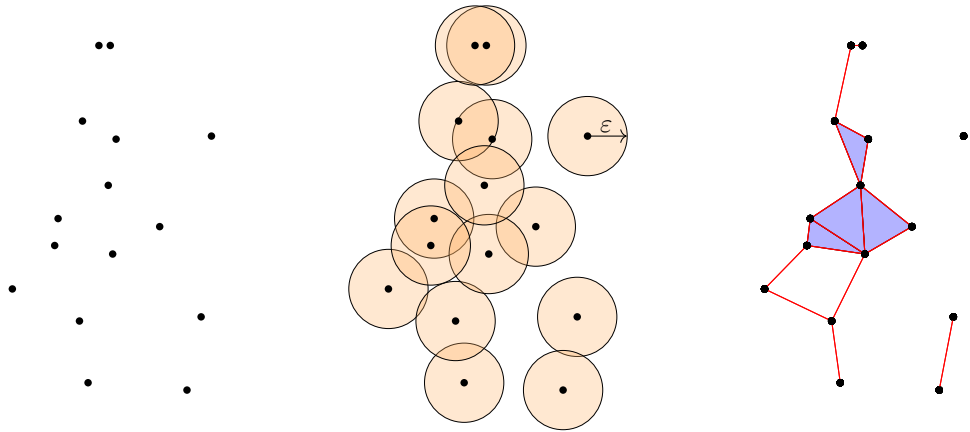
\begin{figure}[H]
    \centering
    \begin{tikzpicture}[ele/.style={fill=black,circle,minimum width=2pt,inner sep=1pt}, scale = 1]

    \node[ele] (a) at (2.8209087137275, 1.18093727264829) {};
    \node[ele] (b) at (2.95833231728538,	3.58594837328201) {};
    \node[ele] (c) at (0.311441915093690,	1.55228774215654) {};
    \node[ele] (d) at (1.69060866126193,	3.54516924716448) {};
    \node[ele] (e) at (1.31799287927655,	0.304557156170160) {};
    \node[ele] (f) at (1.61347687509477,	4.79163906483583) {};
    \node[ele] (g) at (2.27043233468693,	2.38135190156370) {};
    \node[ele] (h) at (1.20381702801476,	1.12542610389225) {};
    \node[ele] (i) at (0.919304590945695,	2.48862221210784) {};
    \node[ele] (j) at (1.64512903092519,	2.01677357394698) {};
    \node[ele] (k) at (1.24383442716840,	3.78580579686650) {};
    \node[ele] (l) at (1.58646607060451,	2.93089871716034) {};
    \node[ele] (m) at (0.874518089942222,	2.13017381291781) {};
    \node[ele] (n) at (1.46214937616206,	4.79124963787566) {};
    \node[ele] (o) at (2.63312304723148,	0.206799557528458) {};

\begin{scope}[elem/.style={circle, draw = black, line width = 0.2pt, fill = orange!60, fill opacity=0.3, minimum width = 30pt}, shift={(5,0)}]

\node[elem] (a) at (2.8209087137275, 1.18093727264829) {};
\node[elem] (b) at (2.95833231728538,   3.58594837328201) {};
\node[elem] (c) at (0.311441915093690,  1.55228774215654) {};
\node[elem] (d) at (1.69060866126193,   3.54516924716448) {};
\node[elem] (e) at (1.31799287927655,   0.304557156170160) {};
\node[elem] (f) at (1.61347687509477,   4.79163906483583) {};
\node[elem] (g) at (2.27043233468693,   2.38135190156370) {};
\node[elem] (h) at (1.20381702801476,   1.12542610389225) {};
\node[elem] (i) at (0.919304590945695,  2.48862221210784) {};
\node[elem] (j) at (1.64512903092519,   2.01677357394698) {};
\node[elem] (k) at (1.24383442716840,   3.78580579686650) {};
\node[elem] (l) at (1.58646607060451,   2.93089871716034) {};
\node[elem] (m) at (0.874518089942222,  2.13017381291781) {};
\node[elem] (n) at (1.46214937616206,   4.79124963787566) {};
\node[elem] (o) at (2.63312304723148,   0.206799557528458) {};

\node[ele] (a1) at (2.8209087137275, 1.18093727264829) {};
\node[ele] (b1) at (2.95833231728538,    3.58594837328201) {};
\node[ele] (c1) at (0.311441915093690,   1.55228774215654) {};
\node[ele] (d1) at (1.69060866126193,    3.54516924716448) {};
\node[ele] (e1) at (1.31799287927655,    0.304557156170160) {};
\node[ele] (f1) at (1.61347687509477,    4.79163906483583) {};
\node[ele] (g1) at (2.27043233468693,    2.38135190156370) {};
\node[ele] (h1) at (1.20381702801476,    1.12542610389225) {};
\node[ele] (i1) at (0.919304590945695,   2.48862221210784) {};
\node[ele] (j1) at (1.64512903092519,    2.01677357394698) {};
\node[ele] (k1) at (1.24383442716840,    3.78580579686650) {};
\node[ele] (l1) at (1.58646607060451,    2.93089871716034) {};
\node[ele] (m1) at (0.874518089942222,   2.13017381291781) {};
\node[ele] (n1) at (1.46214937616206,    4.79124963787566) {};
\node[ele] (o1) at (2.63312304723148,    0.206799557528458) {};

\node (epedge) at (3.48, 3.58594837328201) {};
\node (ep) at (3.21, 3.71) {$\varepsilon$};
\draw[->] (b.center) -- (epedge.center) ;

\end{scope}

\begin{scope}[shift={(10,0)}]
    \node[ele] (a) at (2.8209087137275, 1.18093727264829) {};
    \node[ele] (b) at (2.95833231728538,	3.58594837328201) {};
    \node[ele] (c) at (0.311441915093690,	1.55228774215654) {};
    \node[ele] (d) at (1.69060866126193,	3.54516924716448) {};
    \node[ele] (e) at (1.31799287927655,	0.304557156170160) {};
    \node[ele] (f) at (1.61347687509477,	4.79163906483583) {};
    \node[ele] (g) at (2.27043233468693,	2.38135190156370) {};
    \node[ele] (h) at (1.20381702801476,	1.12542610389225) {};
    \node[ele] (i) at (0.919304590945695,	2.48862221210784) {};
    \node[ele] (j) at (1.64512903092519,	2.01677357394698) {};
    \node[ele] (k) at (1.24383442716840,	3.78580579686650) {};
    \node[ele] (l) at (1.58646607060451,	2.93089871716034) {};
    \node[ele] (m) at (0.874518089942222,	2.13017381291781) {};
    \node[ele] (n) at (1.46214937616206,	4.79124963787566) {};
    \node[ele] (o) at (2.63312304723148,	0.206799557528458) {};
    \foreach \firstnode in {a, b, c, d, e, f, g, h, i, j, k, l, m, n, o}{%
   \foreach \secondnode in {a, b, c, d, e, f, g, h, i, j, k, l, m, n, o}{%
   \setveclength{\mydist}{\firstnode}{\secondnode}
   \pgfmathparse{\mydist < 30 ? int(1) : int(0)}
   \ifnum\pgfmathresult=1
     \draw[red] (\firstnode.center) -- (\secondnode.center);
   \fi
  }
    \node[ele] (a) at (2.8209087137275, 1.18093727264829) {};
    \node[ele] (b) at (2.95833231728538,	3.58594837328201) {};
    \node[ele] (c) at (0.311441915093690,	1.55228774215654) {};
    \node[ele] (d) at (1.69060866126193,	3.54516924716448) {};
    \node[ele] (e) at (1.31799287927655,	0.304557156170160) {};
    \node[ele] (f) at (1.61347687509477,	4.79163906483583) {};
    \node[ele] (g) at (2.27043233468693,	2.38135190156370) {};
    \node[ele] (h) at (1.20381702801476,	1.12542610389225) {};
    \node[ele] (i) at (0.919304590945695,	2.48862221210784) {};
    \node[ele] (j) at (1.64512903092519,	2.01677357394698) {};
    \node[ele] (k) at (1.24383442716840,	3.78580579686650) {};
    \node[ele] (l) at (1.58646607060451,	2.93089871716034) {};
    \node[ele] (m) at (0.874518089942222,	2.13017381291781) {};
    \node[ele] (n) at (1.46214937616206,	4.79124963787566) {};
    \node[ele] (o) at (2.63312304723148,	0.206799557528458) {};
}
\begin{pgfonlayer}{background}
\path[fill=blue!30, draw] (g.center) to (j.center) to (l.center) to (g.center); 
\path[fill=blue!30, draw] (d.center) to (k.center) to (l.center) to (d.center); 
\path[fill=blue!30, draw] (m.center) to (i.center) to (j.center) to (m.center); 
\path[fill=blue!30, draw] (l.center) to (j.center) to (i.center) to (l.center); 
\end{pgfonlayer}

\end{scope}

\end{tikzpicture}
    \caption{The process of constructing a VR complex.}
    \label{fig: Rips}
\end{figure}
This process is depicted in Fig. \ref{fig: Rips} for some randomly-sampled data and an arbitrary value of $\varepsilon$, which can be seen as the radius of the balls. The existence of a simplex is determined by how the balls intersect between the vertices. For a VR complex, a simplex between some set of vertices is formed if the Euclidean distance between all those vertices is less than $\varepsilon$.

\subsection{Persistent Homology}
\label{sec:persisthom}
Before delving into persistent homology, the homology of a topological space needs to be defined. Many great introductions and summaries of homology already exist; the more intrigued reader may want to consult \cite{genomics, ghrist2018homological, nash1988topology, boissonnat2018geometric, schutz1980geometrical, ghrist2014EAT}. In very quick terms, the homology is calculated by computing successive boundary operations on a \textit{chain complex} of a topological space. Homology is essentially the quantification of the number of voids present in a topological space, or in the case of TDA, a simplicial complex. The homology groups, $H_k(X)$ are invariants for the data set $X$, where $k$ refers to the dimension of the homology group. Generally, the $k^{\text{th}}$ homology group encodes information about the number of $k-$dimensional holes in the data. Under the rules of topology, discontinuities (voids) cannot be created or destroyed under continuous maps (homeomorphisms). Therefore, a simplicial complex can be categorised by the properties underpinned by the homology. The homology can be used to categorise and compare between simplicial complexes, and by extension, data sets.

From the homology, the \textit{Betti numbers} are defined as the \textit{rank} of the homology groups. If the Betti numbers for two topological spaces are different, these spaces are not topologically identical, meaning a continuous bijective map between the spaces does not exist. The zeroth Betti number, $\beta_0$, is the rank of the zeroth homology group, $H_0(X)$, and refers to the number of connected sets in the topological space. The first Betti number, $\beta_1$, is the rank of the first homology group, $H_1(X)$, and refers to the number of non-contractible holes present in the space. The second Betti number, $\beta_2$, refers to the number of enclosed volumes in the topological space. This analogy carries on further for higher dimensions.

This discussion now raises the question: which length scale $\varepsilon$ is representative of the topology of the data? When constructing the VR complexes, for the same data set, different values of epsilon, will result in different values for the Betti numbers. The hyper-parameter $\varepsilon$ determines the Betti numbers for that specific instance of some point cloud data. Additionally, when the feature present within the data is at a length scale less than $\varepsilon$ this feature will not be expressed, as $\varepsilon$ will span the feature. This reasoning means that only topological properties that are described at a length scale greater than $\varepsilon$ can be captured. A problem arises here, as usually the feature scale is not known prior to analysis, and a manifold may have many multi-scale features. 

\begin{figure}[H]
    \centering
    \includegraphics[width = 0.8\textwidth]{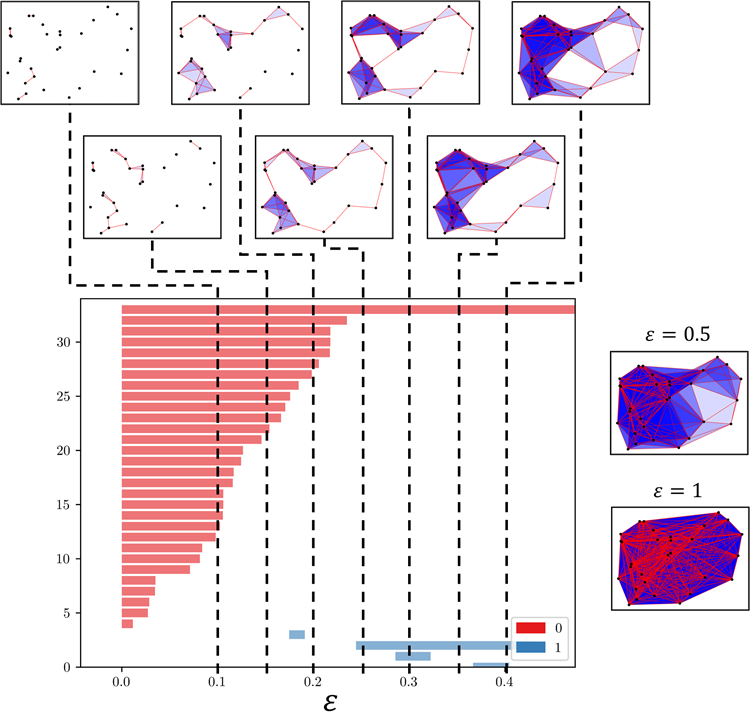}
    \caption{A persistence barcode with realisations showing which simplicial complexes are present at values of $\varepsilon$ of $0.10$, $0.15$, $0.20$, $0.25$, $0.30$, $0.35$, $0.40$, $0.50$, and $1.00$.}
    \label{fig:persistencerealisations}
\end{figure}

The answer to this problem is to vary $\varepsilon$ and see how the Betti numbers evolve and \textit{persist}. When varying $\varepsilon$, a \textit{filtered simplicial complex} is formed. This can be thought of as a chain or sequence of simplices; with $n$ disconnected points at the beginning and a fully connected $(n-1)$- simplex at the end. The next simplicial complex in the \textit{filtration} is the previous simplicial complex plus the next simplex to be formed as $\varepsilon$ increases. Fig. \ref{fig:persistencerealisations} shows some simplicial complexes in this process. A filtered simplicial complex is an ordering of simplices to show how they evolve as a distance scale is increased. From this, one can form an idea of how the topology evolves over the filtration. 

Obtaining the homology for a single value of $\varepsilon$ provides very limited information; this notion is almost redundant, because of potential varying feature length scales in the manifold. For this reason, it is vital to consider what homological features persist as $\varepsilon$ is varied. The goal of persistent homology is to track the homology classes as $\varepsilon$ is varied. The process of varying $\varepsilon$ does not bias any disk size, as all are being considered. This process will give an initial value, $\varepsilon_{\text{min}}$, where a specific homological feature comes to life and $\varepsilon_{\text{max}}$, where the feature is no longer considered for that simplicial complex. This range of values $[\varepsilon_{\text{min}},\varepsilon_{\text{max}}]$ is called the \textit{persistence interval} for that homological feature. Each persistence interval is attributed a Betti number. Following this, the set of all persistence intervals is descriptive for that manifold, giving information about in which dimension a hole exists in the data and over what range of values it persists. 

The persistence intervals obtained can be represented in two ways: \textit{barcodes} or \textit{persistence diagrams}. In a barcode representation, the $x-$axis refers to the value of $\varepsilon$. As $\varepsilon$ increases, the barcode shows which features persist. The set of intervals are plotted with each interval beginning at its $\varepsilon_{\text{min}}$ and ending at its $\varepsilon_{\text{max}}$. The colour of the interval on the barcode refers to  the Betti number, $\beta_k$ \cite{ghrist2008barcodes}. The value of the $y-$axis can simply be thought of as an indexing of the intervals in the barcode. 

An example of a barcode can be seen in Fig. \ref{fig:persistencerealisations}, with vertical dotted lines showing the intersections with the intervals, showing which features are present for the corresponding simplicial complex in the filtration. The red bars refer to the persistence intervals for $\beta_0$, similarly the blue bars are for $\beta_1$. A bar is formed when that feature begins and ends when that feature dies. The number of intersections of the persistence intervals with the slices (denoted by the dashed black line) at each $\varepsilon$ refers to the Betti number for each feature. For instance, for $\varepsilon = 0.3$, the simplicial complex consists of one connected component (as there is only one intersection with the red persistence intervals) and there are two holes present (denoted by two intersections with the blue persistence intervals). For $\varepsilon=0.50, 1.00$, these are shown besides the barcode, as there is no change in the barcode for these values. For $\varepsilon>0.45$, all of the holes have been spanned, and there is only one connected component. The vertices only become more and more connected as $\varepsilon$ increases, and this has little topological interest. The length of the interval represents for how long the feature persists. The longer the feature persists, the higher the probability that this feature is characteristic of the manifold. Shorter intervals are generally regarded as topological noise.

\begin{figure}[H]
    \centering
    \includegraphics[width = 0.6\textwidth]{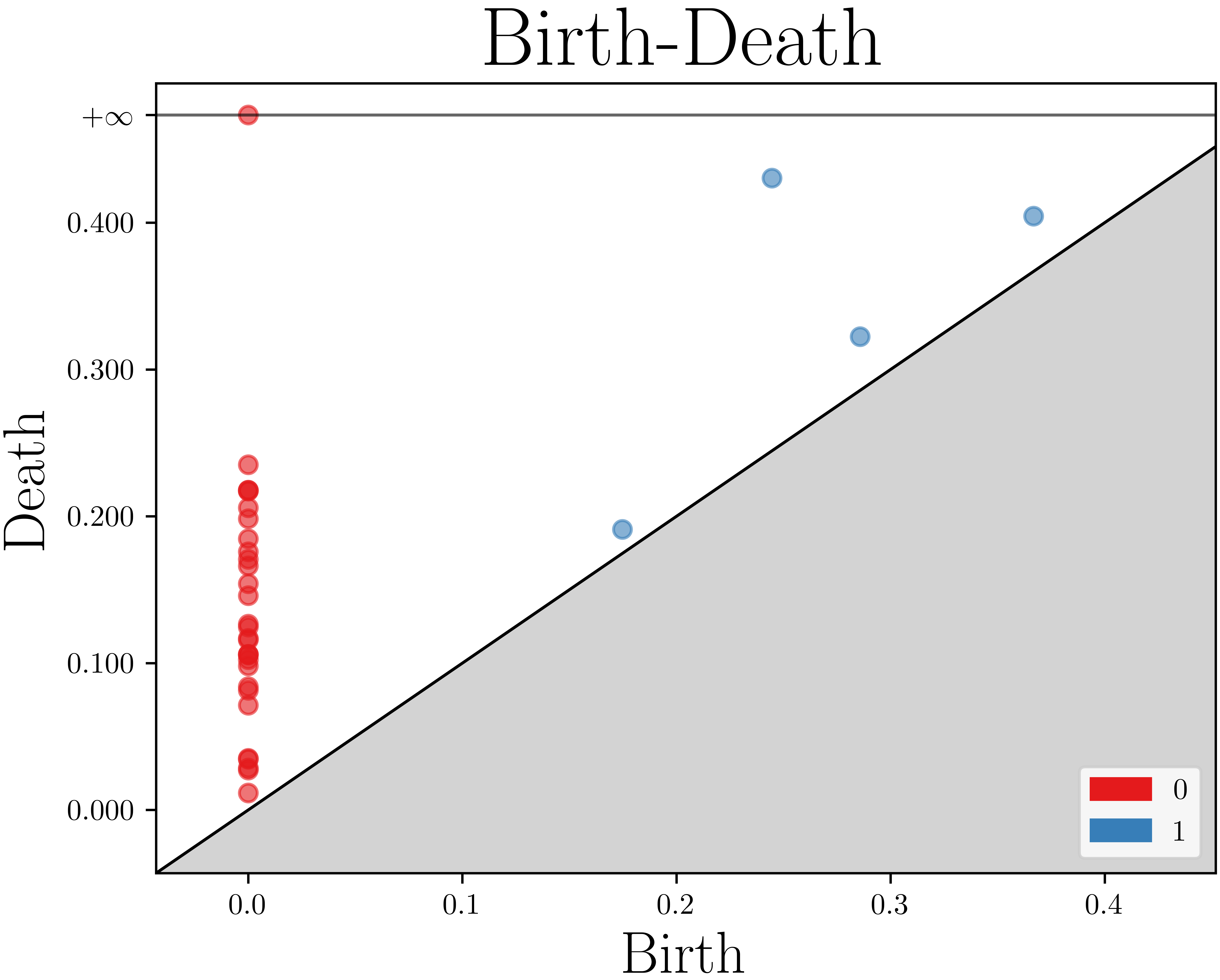}
    \caption{Birth-death diagram for the same random data present in Fig. \ref{fig:persistencerealisations}.}
    \label{fig:BDdia}
\end{figure}

The other method of visualising the set of persistence intervals is the persistence diagram, or \textit{birth-death} diagram. In this representation, $\varepsilon_{\text{min}}$ is plotted on the $x-$axis and $\varepsilon_{\text{max}}$ is plotted on the $y-$axis, with each interval represented by the point $(\varepsilon_{\text{min}},\varepsilon_{\text{max}})$. Intuitively, there is a line defined by $y=x$, below which points will not be plotted. This fact means that the grey region in Fig. \ref{fig:persistencerealisations} will never contain any points. The line $y=x$ has the interpretation that the feature must first exist before it can die. Reading these diagrams is as intuitive as reading the barcodes; the vertical height of the point from the line $y=x$ is analogous to the length of the interval, that is, the further a point is from the line $y=x$, the more the feature is persistent. An example of a birth-death diagram can be seen in Fig. \ref{fig:BDdia}.

On both the barcode and birth-death diagrams, it can be seen that one feature persists to $\infty$; this is because there will always be a fully-connected simplex that persists to infinity. There will be a value of $\varepsilon_{\text{fc}}$ that results in a fully-connected simplex, where every vertex is connected to every other vertex. For values $\varepsilon>\varepsilon_{\text{fc}}$ the simplex will remain fully connected, and therefore this will continue to infinity. The removal of this infinite interval is called the \textit{reduced homology}. The reduced homology is required for use in many calculations.

The space of barcodes actually forms a metric space \cite{genomics}; the distance between the barcodes is a measure of similarity of two barcodes. As the persistent intervals are invariant for a manifold, the data manifolds can be represented by their persistent homology. This notion of a metric space allows one to compare the similarity of manifolds. Metrics between barcodes are well established and the one used in this paper is the \textit{$p-$Wasserstein distance} \cite{lacombe2018large}.

\begin{defn}
Given two barcodes $B_1$ and $B_2$. For $p>0$, the \textit{$p-$Wasserstein distance}, $\partial_{W_p}$, is given by,
$$\partial_{W_p}(B_1,B_2) = \left(\text{inf} \sum_{Z\in B_1}d_\infty(Z, \phi(Z))^p \right)^{\frac{1}{p}}$$
where $\phi$ is a matching between $B_1$ and $B_2$, $d_\infty$ is the supremum metric, and $Z$ is a persistence interval in $B_1$ \cite{genomics}.
\end{defn}

The Wasserstein distance is used to measure the similarity between two persistence barcodes, by computing the sum over their edge lengths. The q-Wasserstein distance is defined as the minimal value achieved by a perfect matching between the points of the two diagrams, where the value of a matching is defined as the $q^{\text{th}}$ root of the sum of all edge lengths to the power $q$ \cite{lacombe2018large}.

\subsection{Calculating Fractal Dimension}
The first application of persistent homology in this paper is on calculating \textit{fractal dimension}, which is a measure of how much space a set occupies. To gauge the idea of the fractal dimension intuitively, consider that a single point is zero-dimensional, a line spanning two points is one-dimension, and the area enclosed by three points is two-dimensional, and so on. Now, suppose a curve, consisting of purely one-dimensional elements has infinite length and is bounded by a finite region. This curve may be believed to occupy more space than something that is only considered one-dimensional, but it also does not completely fill this arbitrarily-defined space. A logical conclusion is that the dimension of this curve is somewhere in \textit{between} one and two, depending on how efficiently it occupies this area.

The first notable work on quantifying fractal dimension was by Hausdorff \cite{hausdorff1918dimension}, where a measure of roughness was conceived; this was later expanded on by Falconer, and generalised for point clustering in \cite{falconer1986geometry}. Mandelbrot, popularised the idea of fractal dimension in the notable work \cite{mandelbrot1982fractal}.

The two most common methods of quantifying fractal dimension are the \textit{Hausdorff} and \textit{box-counting} dimensions. The box-counting method exploits the dimensional relations to scalability and the difference in occupied space over different scales. The box-counting method is very popular because of its ease of computation. The method of using the persistent intervals, which is described later, more closely represents the box-counting method.

To calculate the box-counting dimension, assume a space $X$. The smallest number of sets of diameter $\epsilon$ that can cover $X$, is referred to as $N(\epsilon)$. The scale $\epsilon$ determines $N(\epsilon)$, and the dimension of $X$ determines the rate of change of $N(\epsilon)$. As the scale is altered, it is expected that,

\begin{equation}
    N(\epsilon) \simeq c\epsilon^{-d_B}
\end{equation}

for positive constants $d_B$ and $c$. It is said that $X$ has a box dimension of $d_B$. To solve for the dimension of the set, this equation can be rearranged to give,
\begin{equation}
    d_B = \lim_{\epsilon \rightarrow 0} \frac{\log(N(\epsilon))}{-\log(\epsilon)},
\end{equation}

with the constant term, $c$, disappearing in the limit \cite{falconer1986geometry}.

Inherently, fractals are infinitely-complex objects that are self-similar, often over an infinite range of scales. This view gives a problem when working with finite point-data sets, as only finite information can be captured in a finite point-data set. When zooming in excessively, points will become more sparse and the approximations will inevitably become less exact. In reality for finite sets, self similarity may only be visible over a small number of scales. This process can be seen later, in Fig. \ref{fig: Henonzooms}, where the self similarity is clear over two length scales, then thereafter points become sparse.

\subsection{Fractal Dimension - Persistent Homology}
Persistent homology can be used to calculate fractal dimension, given some point data sampled from a fractal. This idea first came from that of the \textit{minimal spanning tree} (MST) \cite{kruskal1956shortest, prim1957shortest}. The MST method was shown as a viable method to calculate the fractal dimension of a set \cite{van1992minimal}. Numerous works have showed that the MST method is equivalent to using the zeroth homology group for calculating the fractal dimension \cite{schweinhart2020fractal, schweinhart2021persistent, van1992minimal, kozma2006minimal}. Unlike using the MST method, the persistent homology can be generalised for higher-order homology groups, therefore returning more information regarding the topology of the fractal shapes.

Often when analysing the persistent data, the smaller intervals are discarded as noise, as these mean that this specific feature only persists for a short while. In the case of calculating fractal dimension, there is information contained in all of the persistent intervals, as the idea here is to deduce the fractal dimension of $X_n$ points as $n \rightarrow \infty$ and seeing how the lengths of the persistence intervals in each homology group vary. The shorter intervals provide a good measure of how the local geometry is present in the finite random sample.

To calculate the fractal dimension from the persistent homology, the $\alpha$-weighted sum for $\alpha > 0$, of the persistent intervals, in a given dimension should be computed, as follows,
\begin{equation}
    E^i_\alpha(x_1,...,x_n) = \sum_{I \in PH_i(x_1,...,x_n)} |I|^\alpha,
    \label{eq: PH cumsum}
\end{equation}

$(x_1,...,x_n)$ are $n$ points sampled from a manifold, most interestingly one that exhibits fractal behaviour. $PH_i(x_1,...,x_n)$ is the $i^{\text{th}}$ dimension persistent homology group. $|I|$ is the length of the interval from $PH_i$. The alpha-weighted sum of the persistence intervals tracks the rate at which the topological noise decays, and similarly to the box-counting method, this can be linked to the fractal dimension \cite{jaquette2020fractal}. The value of alpha will give a larger weighting to longer intervals.

\begin{defn}
Let $X$ be a bounded subset of a metric space and $\mu$ a measure on $X$. For all $i \in \mathbb{N}$ and a value $\alpha > 0$, which is used to give a weighting to larger persistence intervals in the calculation \cite{jaquette2020fractal}, the persistent homology dimension can be defined as,
\begin{equation}
    \dim_{PH_i^\alpha}(\mu) = \frac{\alpha}{1-\beta},
\end{equation}
where,
\begin{equation}
    \beta = \lim_{n \rightarrow \infty} \sup \frac{\log (\mathbb{E}(E^i_\alpha (x_1,...,x_n)))}{\log(n)}
\end{equation}
Here, $\sup$ refers to the largest value in the set. The operator $\mathbb{E}$, is used as the expected value of a random variable. Finally, $E^i_\alpha(x_1,...,x_n)$ is the alpha-weighted cumulative sum over the $n$ points. This result means that the embedding dimension $d = \dim_{PH_i^\alpha}(\mu)$, of a manifold can be calculated if $E_\alpha ^i(x_1,..., x_n)$ scales with $n^{\frac{d-\alpha}{d}}$ \cite{schweinhart2021persistent, jaquette2020fractal}.
\end{defn}

For the case when $i=0$, the persistence intervals are equivalent to the lengths of the edges in the MST. From the MST, there is a set of $n$ vertices and a set of $n-1$ edges, where each edge spans two vertices. This work is built on ideas from Kruskal \cite{kruskal1956shortest} and Prim \cite{prim1957shortest} in the 1950s. The edges in the MST are equivalent to the persistence intervals in the reduced zeroth homology group.

The MST approach formulates the problem in terms of graph theory, with $V$ being the set of \textit{vertices} and $E$ being the \textit{edge-set}, connecting the vertices. Two vertices are referred to as \textit{connected}, if a \textit{path} connects them. A \textit{path} is the successive joining of adjacent edges from one vertex to another, i.e, one can walk from one vertex to the other without leaving the path. If these edges form a closed loop, this is called a \textit{circuit}. If a graph does not contain a closed circuit, it is called a \textit{tree}.

\begin{figure}[H]
    \centering
    \subfloat[{VR Complex}]{\includegraphics[width = 0.45\textwidth]{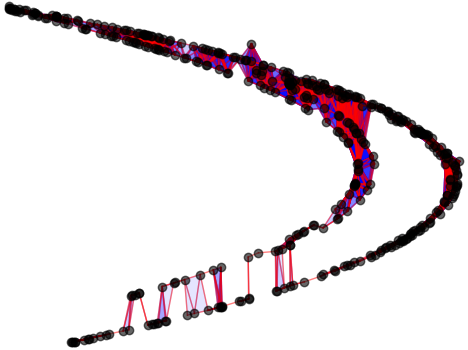}}
    \qquad
    \subfloat[{MST}]{\includegraphics[width = 0.45\textwidth]{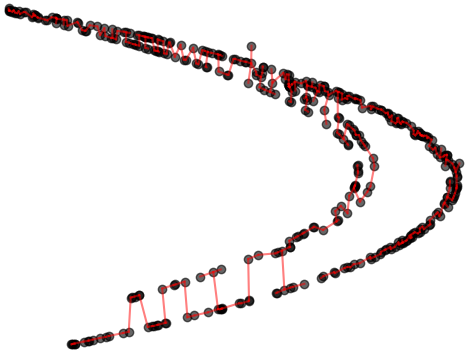}}
    \qquad
    \caption{Hénon attractor constructed with the two methods of calculating fractal dimension.}
    \label{fig: randomMST}
\end{figure}

A spanning tree has an attribute called its \textit{length}. The length of a spanning tree is the sum of all the edges in the tree. The MST is the spanning tree that spans the points most efficiently, by minimising the length of the edges. The algorithms used to calculate MSTs are very well optimised, meaning that competitive results for calculating fractal dimensions can be obtained from their usage \cite{jaquette2020fractal}. One can use the lengths of the edges in an MST to approximate the dimension of the manifold from which the vertices are sampled. The asymptotic behaviour of Eq. (\ref{eq: PH cumsum}) is studied to calculate the fractal dimension.

This property is useful, as algorithms for calculating MSTs are much faster than calculating persistent intervals. However, the MST is only equivalent to the zeroth homology group. If more information is required for higher-dimensional homological features, the slower, but more informative persistence algorithms must be used. Fig. \ref{fig: randomMST} shows the VR complex and the MST for an attractor.

When dealing with higher-order persistent homology, $i \geq 1$, things become more tricky. With $PH_0$, it is known that there are going to be $n-1$ edges for $n$ points. Whereas, for higher-order persistent homology, this is not the case. There exist metric spaces where the number of persistent intervals grows faster than linearly in the number of points. To get around this problem, a limit for the upper bound can be proven. In the case of the VR complex, Schweinhart proved $|PH_1| = O(n)$ for the points $(x_1,...,x_n)$ \cite{schweinhart2021persistent}. 

\section{Attractors}
\label{sec: Attractors}
Strange attractors arise from nonlinear dynamical systems. A dynamical system is one which is modelled (and evolves) in a phase space embedded in $\mathbb{R}^n$, where the geometric object is embedded in Euclidean space and parameterised by time.  Attractors represent the state to which a dissipative dynamical system will eventually converge, seemingly regardless of initial conditions. Attractors are common-place for study as they describe the asymptotic behaviour of many dynamical systems \cite{lorenz1963deterministic, rossler1976equation, henon1976two}.

\subsection{Lorenz Attractor}
\label{sec: Lorenz}
The Lorenz attractor is defined by the differential equations,
\begin{align}
    \label{eqnarr: Lorenz1} \dot{x} &= -\sigma \\
    \label{eqnarr: Lorenz2} \dot{y} &= \rho u - v - uw \\
    \label{eqnarr: Lorenz3} \dot{z} &= -\beta w +uv
\end{align}

The attractor was first discovered by Lorenz, when studying non-periodic turbulent flows \cite{lorenz1963deterministic}. It displays an interesting topology, with two holes being present in the manifold. It also shows a Cantor-set like behaviour over its cross sections \cite{viswanath2004fractal} for certain parameters.

\begin{figure}[H]
    \centering
    \subfloat[{Dense Lorenz Attractor.}]{\includegraphics[width = 0.5\textwidth]{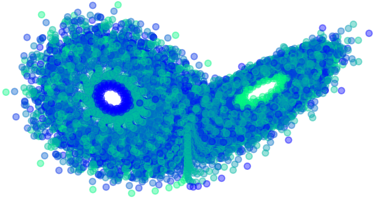}}
    \qquad
    \subfloat[{log-log plot to determine fractal dimension, with gradient (red).}]{\includegraphics[width = 0.45\textwidth]{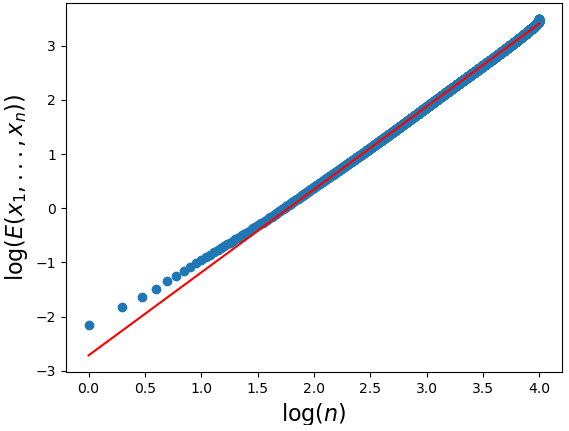}}
    \qquad
    \caption{Lorenz Attractor.}
    \label{fig: Lorenzdim}
\end{figure}
For the system parameters $\rho=28$, $\sigma=10$, and $\beta=\frac{8}{3}$, the fractal dimension has previously been calculated by Viswanath \cite{viswanath2004fractal} to be $2.063$ . To obtain a structure with the fine details, showing the Cantor-like fractal cross sections, a large number of points are required for the embedding. 10000 points were calculated here for a dense embedding of the Lorenz attractor. Using the MST method of calculating the fractal dimension \cite{jaquette2020fractal, 2010march, 2018curtin}, of the embedded attractor shown in Fig. \ref{fig: Lorenzdim}, an approximation to the Hausdorff dimension of $2.0826 \pm 0.03603$ was calculated. This value is calculated by determining the linear gradient of the log-log plot of the cumulative sum of the MST edges, which are analogous to the persistent intervals in $PH_0$. 

\begin{figure}[H]
    \centering
    \subfloat[{Lorenz Attractor.}]{\includegraphics[width = 0.5\textwidth]{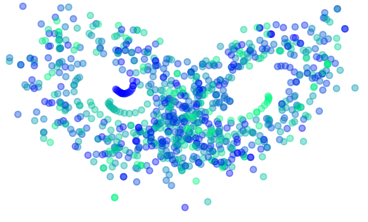}}
    \qquad
    \subfloat[{Persistence Homology for the Lorenz attractor.}]{\includegraphics[width = 0.45\textwidth]{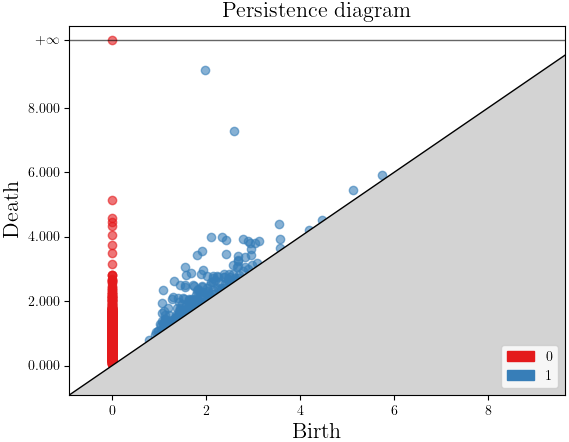}}
    \qquad
    \caption{Lorenz Attractor.}
    \label{fig: Lorenz}
\end{figure}

Now, in order to capture the global topology of the system, fewer points are required to calculate the persistent homology. In this example, a more sparse Lorenz attractor is sampled with only 1000 points, shown in Fig. \ref{fig: Lorenz}. These samples were taken over the same range as the previous case but now there is a greater time step between each point. This procedure ensures a relatively dispersed distribution of points over the Lorenz attractor. By taking a dispersed distribution of points on the manifold, the global topology of the manifold should become more clear.

The birth-death diagram can be seen on the right hand side of Fig. \ref{fig: Lorenz}. There are no large differences in $\beta_0$, represented by the red points; this strongly indicates that the Lorenz attractor consists of only one connected component. For the case of the first homology group, there is a large amount of noise present close to the line $y=x$. Most interestingly from this plot, two features persist long enough to be deemed properties of the manifold. These points are (2.59, 7.29) and (1.97, 9.17). The holes represented by these points can be visibly seen on the left hand side of Fig. \ref{fig: Lorenz} as the holes present within the manifold. The smaller of the two intervals represents the void in the left-hand side of the plot.

\subsection{Hénon Attractor}
\label{sec: henon}
The Hénon attractor is a two-dimensional quadratic map with a constant Jacobian; it was first conceived as a simplified discrete map of the Lorenz system. As the map is discrete, it has become a common object for study in dynamical systems, as computations for generating a large number of points are fast \cite{henon1976two}. The Hénon attractor used in this work is defined by,

\begin{align}
    \label{eqnarr: Henon1} x_{n+1} &= 1 - ax_n^2 + y_n  \\
    \label{eqnarr: Henon2} y_{n+1} &= bx_n 
\end{align}
In this case, the well studied parameters of $a=1.4$, $b=0.3$, and an initial point of $(0.1, 0.3)$  were used, as these give a convergent solution. $2000$ iterations were taken to give a finite approximation of the orbit of this specific Hénon attractor, with the initial conditions listed. Using the MST method of calculating the fractal dimension \cite{jaquette2020fractal, 2010march, 2018curtin}, a value of $1.2558 \pm 0.04476$ was obtained. A previously-calculated value from work by Grassberger \cite{grassberger2004measuring}, calculated the dimension to be 1.26.

\begin{figure}[H]
    \centering
    \subfloat[{Hénon Attractor generated with 2000 points.}]{\includegraphics[width = 0.4\textwidth]{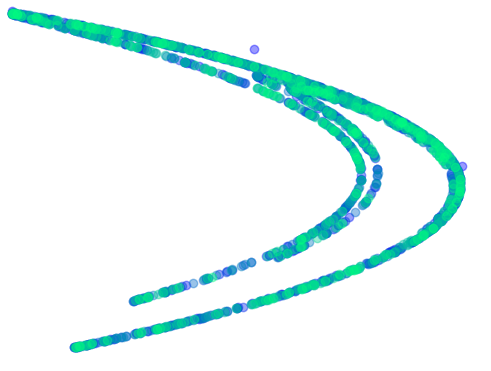}}
    \qquad
    \subfloat[{log-log plot to determine fractal dimension, with gradient (red).}]{\includegraphics[width = 0.45\textwidth]{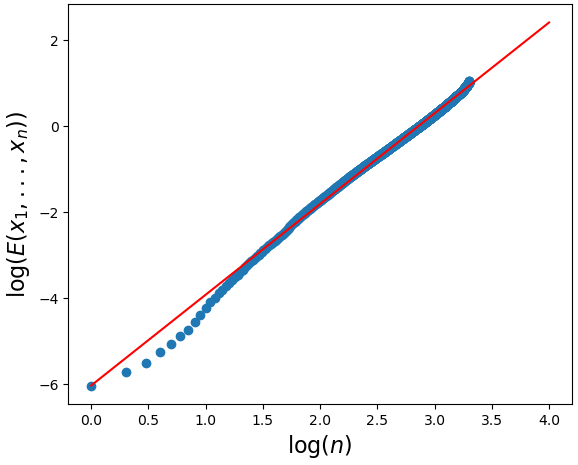}}
    \qquad
    \caption{Hénon Attractor.}
    \label{fig:henon}
\end{figure}

\begin{figure}[H]
    \centering
    \subfloat[{$x\in[0.5, 0.75]$, $y\in[0.15,0.21]$.}]{\includegraphics[width = 0.3\textwidth]{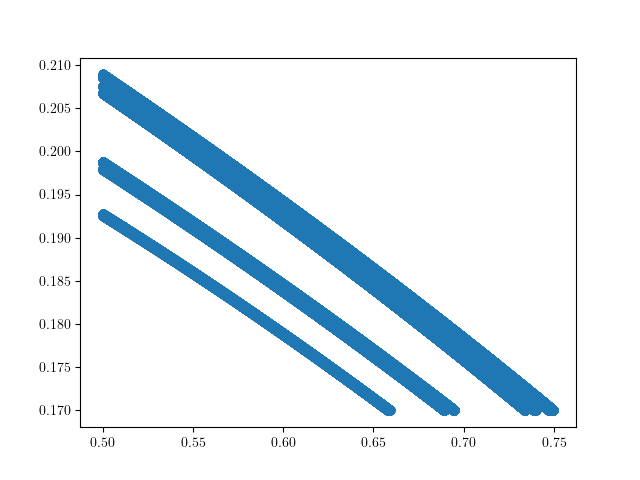}}
    \qquad
    \subfloat[{$x\in[0.62, 0.64]$, $y\in[0.185,0.191]$.}]{\includegraphics[width = 0.3\textwidth]{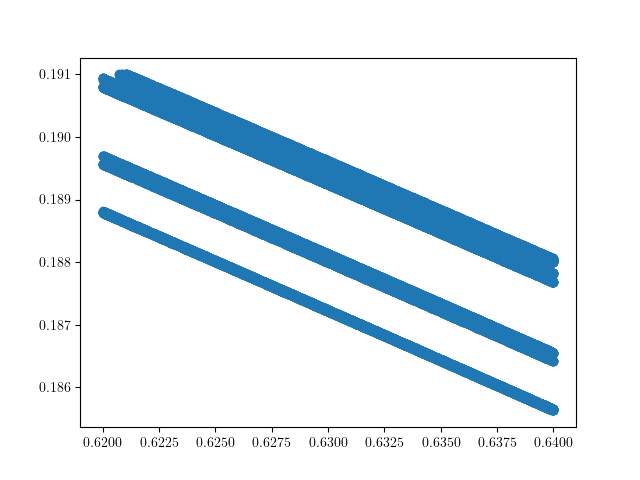}}
    \qquad
    \subfloat[{$x\in[0.63, 0.6325]$, $y\in[0.1889, 0.1895]$.}]{\includegraphics[width = 0.3\textwidth]{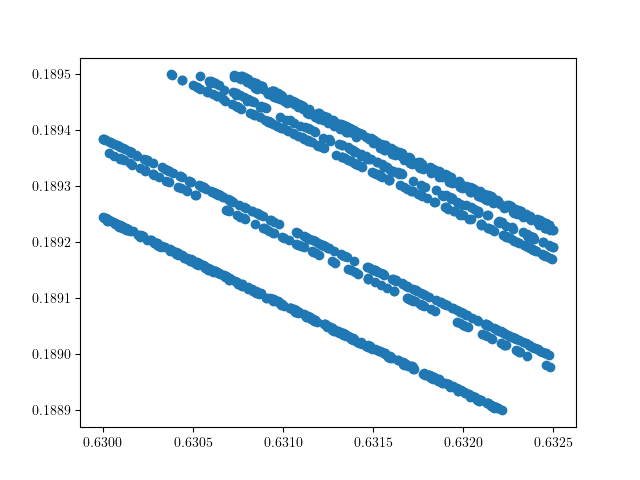}}
    
    \caption{Hénon points sampled from three different scales, showing the Cantor cross section.}
    \label{fig: Henonzooms}
\end{figure}

The Hénon attractor is well known to exhibit a self-similar, Cantor-like, behaviour over its cross sections \cite{henon1976two}. The longitudinal structure of the Hénon attractor is simple, with each curve appearing to be a 1D manifold. The traversal structure is the interesting part; this can been seen to be self similar in Fig. \ref{fig: Henonzooms}.
The dimensions of (a), (b), and (c) in Fig. \ref{fig: Henonzooms} are $1.233$,  $1.245$, $1.526$; this shows that, as the data become more sparse, this fractal dimension calculation method becomes less accurate as more data points are missing at the smaller length scales.

\subsection{Rössler Attractor}
\label{sec: Rossler}

This attractor was first formulated in \cite{rossler1976equation, rossler1979equation}, it is obtained by solving the differential equations,
\begin{align}
    \label{eqnarr: Rossler1} \dot{x} &= -y-z \\
    \label{eqnarr: Rossler2} \dot{y} &= x+ay \\
    \label{eqnarr: Rossler3} \dot{z} &= b + z(x-c)
\end{align}

\begin{figure}[H]
    \centering
    \subfloat[{Rössler Attractor.}]{\includegraphics[width = 0.4\textwidth]{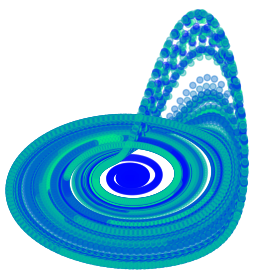}}
    \qquad
    \subfloat[{log-log plot to determine fractal dimension, with gradient (red).}]{\includegraphics[width = 0.45\textwidth]{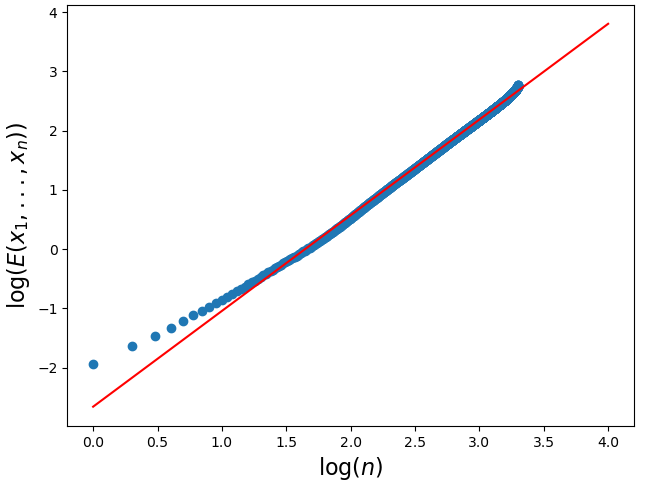}}
    \qquad
    \caption{Rössler Attractor with 20000 points, used to calculate the embedding dimension.}
    \label{fig: Rosslerdim}
\end{figure}

The values for the initial conditions and system parameters were $a=0.2$, $b=0.2$, $c=5.7$, with the initial point, $p_0 = (0,0,0)$. With 20000 points sampled from the Rössler attractor, the dimension estimate gives a value of $2.025 \pm 0.0246$ \cite{jaquette2020fractal, 2010march, 2018curtin}. Kuznetsov calculated the dimension of the Rössler attractor, for these parameters to be $2.0160$ \cite{kuznetsov2018note}. 

\begin{figure}[H]
    \centering
    \subfloat[{Rössler Attractor.}]{\includegraphics[width = 0.4\textwidth]{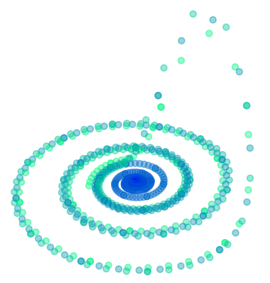}}
    \qquad
    \subfloat[{Persistence Homology of the Rössler attractor.}]{\includegraphics[width = 0.5\textwidth]{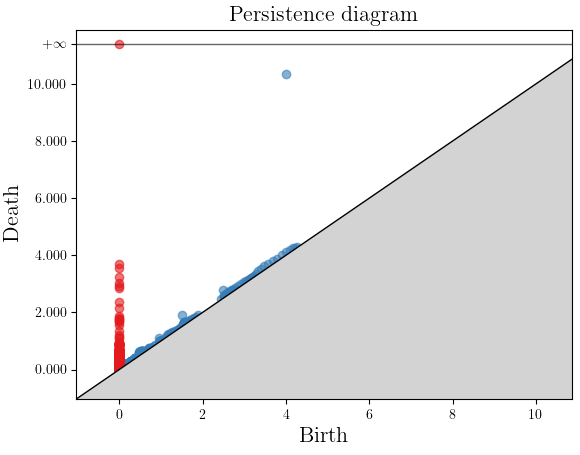}}
    \qquad
    \caption{Rössler Attractor, with 700 points, used to analyse the topology.}
    \label{fig: Rossler}
\end{figure}

From looking at the persistence barcode in Fig. \ref{fig: Rossler}, it can be seen that all the red dots, corresponding to $\beta_0$ are close and show no clear split between these values. This observation strongly implies that the Rössler attractor is a fully-connected manifold. The range in which these occur is most likely because of the data points along the 'flick' of the Rössler attractor being more sparse; if this were a continuous embedding, these features would not be present.

For the case of the first-dimensional Betti numbers, represented by the blue dots in Fig. \ref{fig: Rossler}, there is a lot of topological noise close to the line $y=x$. Interestingly, there is a point far from the line $y=x$, which is representative of the topology of the Rössler attractor. This point has the coordinates $(4.00, 10.3)$, and is representative of the hole formed over the 'flick' of the attractor. These persistence data will now act as the benchmark for the Rössler attractor, and multiple embeddings will be compared to this. The aim here is to minimise the Wasserstein distance (WD) between the reconstructed object and the original attractor. The smaller the WD, the closer the topologies.

\subsubsection{Reconstruction from Time Delay Embedding}
In this case, the topologies of a reconstructed phase space, formed from a 1D time series from the attractor, and the original point cloud are compared. Persistent homology will be used to determine the optimal embedding parameter from the 1D time series.

Time-delay embeddings were first introduced in \cite{packard1980geometry} as a method of inducing geometry from a one-dimensional time series. This work was further expanded on by Takens \cite{takens1981detecting}, proving that the topology can be perfectly reconstructed for chaotic attractors. Skraba et al \cite{skraba2012topological}, then showed that persistent homology could be used with the time-delay embedding to yield useful topological results. 

Given a time-varying series, $f:t \rightarrow \mathbb{R}$, the time-delay embedding can be stacked $d$ times, each with a delay $\alpha$, to give a new embedding $\phi : t \rightarrow \mathbb{R}^d$ where the new embedding is represented by,

\begin{equation}
    \phi(t) = (f(t), f(t+\alpha), ..., f(t+(d-1)\alpha)).
\end{equation}

From this embedding, the time series now has a topology induced. This embedding is a reconstruction of the topology from a one-dimensional time series into any desired dimension. Takens' theorem shows that under certain circumstances, this reconstructed attractor is homeomorphic to the original attractor \cite{takens1981detecting}.

The delay embedding highlights periodic recurrent features in the time series.  Recurrent behaviour will be highlighted in the time-delay embedding as a loop. Persistent homology can then be used to quantify the size of these loops.

If the delay is too small, there will not be sufficient information to form meaningful topology, and the reconstruction will be too similar to a straight diagonal line. Therefore, by maximising the size of the holes present in the attractor by using the persistent homology, an optimal time-delay embedding can be determined. Conversely, using a too-large delay will result in nonsense, as the gap between the readings will be too large and will not show a local change over the manifold, resulting in a deformed attractor.

\begin{figure}[H]
    \centering
    \includegraphics[width= 0.5\textwidth]{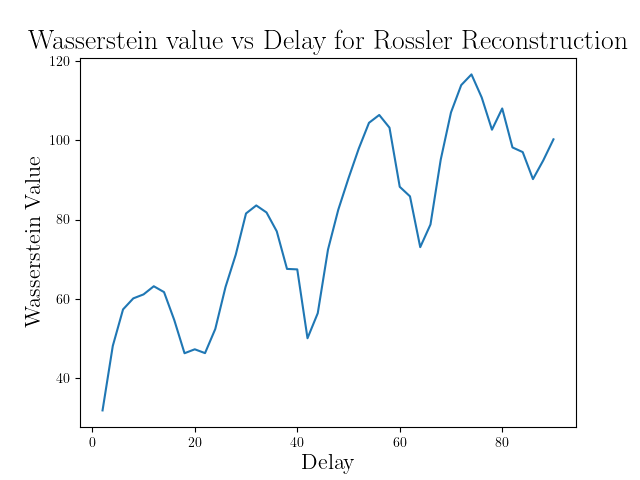}
    \caption{WD vs time delay embedding for Rössler.}
    \label{fig:RosslerwassTDE}
\end{figure}

By computing a range of delay embeddings, the persistent homology of these can be calculated and then the WD can be used as a metric of topological similarity, to give the optimal delay embedding. Fig. \ref{fig:RosslerwassTDE} shows the WD between the original Rössler attractor and attractors formed from a delay embedding over a range of delays. By taking the delay for the minimal WD, this implies that the topology of the original attractor and the reconstructed delay embedding have the most similar topology. Therefore, the delay that minimises the WD is the one that gives the optimal delay embedding. 

There is a clear periodic trend emerging in Fig. \ref{fig:RosslerwassTDE}; this trend occurs as the time-delay embedding comes in and out of phase with the original attractor. The general trend on Fig. \ref{fig:RosslerwassTDE} shows an increase in the WD as the delay is increased. This trend is because of artefacts induced in the reconstruction, as successive points are sampled over different periods from the attractor.

\begin{figure}[H]
    \centering
    \subfloat[{$\alpha = 1$.}]{\includegraphics[width = 0.4\textwidth]{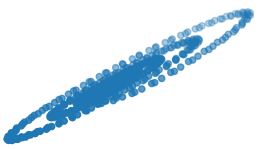}}
    \qquad
    \subfloat[{$\alpha = 57$.}]{\includegraphics[width = 0.4\textwidth]{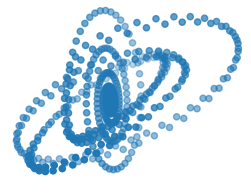}}
    \qquad
    \caption{Examples of Bad Rössler Reconstructions.}
    \label{fig: Rosslerbadembed}
\end{figure}

From Fig. \ref{fig:RosslerwassTDE} it can be seen that a delay equal to one gives the smallest WD; in this case, the reconstructed topology is too similar to the diagonal set, and the topology is relatively uninteresting. If one was to assume an delay embedding of $\alpha=0$, this would result in the straight line $x=y=z$, which will clearly have an uninteresting topology. Therefore, in this analysis, only values after the first peak $(\alpha = 13)$ will be considered, as these ensure an interesting topology is being formed from the embedding.  On the other hand, using a too-large value for the delay gives a deformed manifold. Two bad examples of delay embeddings, $\alpha=1$ and $\alpha=57$, can be seen in Fig. \ref{fig: Rosslerbadembed}. 

\begin{figure}[H]
    \centering
    \subfloat[{ $\alpha=19$.}]{\includegraphics[width = 0.4\textwidth]{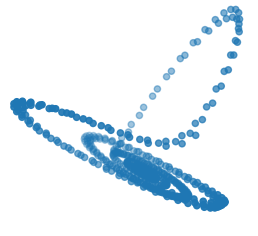}}
    \qquad
    \subfloat[{Persistence Homology.}]{\includegraphics[width = 0.5\textwidth]{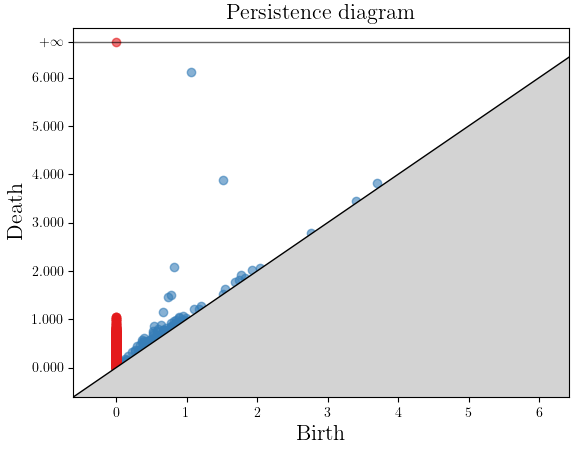}}
    \qquad
    \caption{Optimal Reconstructed Rössler Attractor.}
    \label{fig: Rossleroptimal}
\end{figure}

For $\alpha = 19$, not only the topology of the attractor is most accurately being captured, but from Fig. \ref{fig: Rossleroptimal} (a), there is also a strong likeness to the original geometry. This is not likely to be the case for all other reconstructions, as Takens' only proved this for the topology \cite{takens1981detecting}.

\section{A Case Study of Interest for SHM: the Z24 Bridge}
\subsection{Introduction to the problem}

The Z24 bridge was a structure connecting the two regions, Koppigen and Utzenstorf, in Switzerland. Data were collected over a whole year, with a sensor network placed over the bridge to collect modal parameters. Sensors were also used to measure the air temperature, soil temperature and humidity. Because of the extreme conditions of the Swiss weather, air temperature was recorded as low as $-9^{\circ}C$ and as high as $36^{\circ}C$. As a result, the temperature effects are clearly visible on the calculated natural frequencies \cite{peeters2001one}. Shortly before the destruction of the Z24 bridge, there was controlled damage introduced to the system, which is also visible in the natural frequencies after a certain point in time. 

The change in temperature is the biggest contributing factor to the change in the natural frequency ($\approx 30\%$). The change in temperature has a greater impact than introducing damage ($\approx 7\%$). For this reason, the change in the magnitude of the natural frequencies offers little insight into the presence of damage. The main problem here is to separate the damage case from the temperature effects. 

\begin{figure}
    \centering
    \includegraphics[width= 0.8\textwidth]{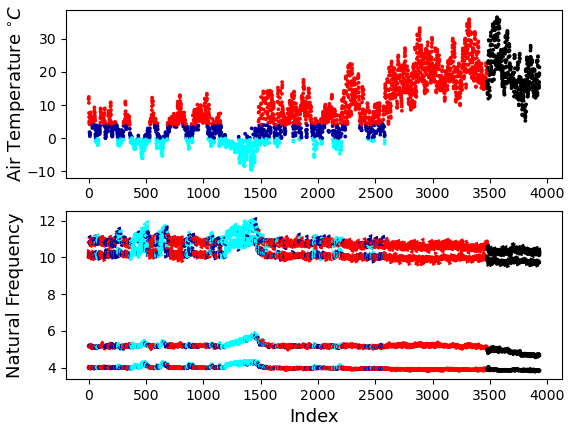}
    \caption{Natural frequency classes according to their temperature.}
    \label{fig:tempnatfreq}
\end{figure}

The data set can be broken down into four categories, according to the air temperature at the time of the measurements, and whether damage was present. Fig. \ref{fig:tempnatfreq} shows the temperature readings and the first four calculated natural frequencies; the corresponding colours refer to: 
\begin{itemize}
    \item Light blue making up the freezing data set. This is any value with a temperature reading $T < 0^{\circ}C$.
    \item Dark blue, making up the cold data set. This is any value in the temperature range $0^{\circ}C \leq T < 4^{\circ}C$.
    \item Red, making up the warm data set. This is any value with a temperature reading $4^{\circ}C \leq T$.
    \item Black, making up the damage-state data set. This is any reading taken after an index of $3475$, irrespective of the temperature.
\end{itemize}

At every measurement instance, the natural frequencies $\{ \omega_i \}$, can be calculated; where $\omega_i$ is the resonance frequency for the $i^{th}$ mode. The first set of the $n$ natural frequencies can be represented as a point in $\mathbb{R}^n$, where the $i^{\text{th}}$ axis is for the value of $\omega_i$. As the Z24 data set has been sampled over different environmental and operational variations (EOVs), it is expected that the resonance frequencies will vary, depending on the environmental variables on the day. The bigger the change in the EOVs, the greater the change in $\omega_i$. As the temperature changes, it is expected that material properties will be affected, therefore resulting in a change in the natural frequency.  

In the case of the Z24 Bridge, by sampling over the time frame of a year, there will be slight changes in humidity, air temperature and soil temperature, that mean that each natural frequency will be slightly different from any other day. As the first four natural frequencies have previously been extracted for each reading \cite{peeters2001one}, each point can be plotted in $R^4$. Plotting these points will trace out a manifold shape that is paramaterised by all of these EOVs. The data points are then assumed to lie on a manifold, representative of this specific bridge. TDA can then be used to form an understanding of what is expected for the shape of the natural frequency manifold for the Z24 data set. 

Since the data have been partitioned into freezing, cold, warm, and damage, TDA  can be used to compare the relative shapes of these manifolds. A significant change in the shape of the manifold could potentially be an indicator of the presence of damage.

\subsection{Wasserstein Distance of Z24 partitions}
There will be three case studies presented here, all on the Z24 data set. Each case study will include a new aspect of analysis to make the analysis more concrete.
\begin{enumerate}
    \item The first case study will compare the manifolds embedded in four dimensional space, where each of the axes corresponds to a natural frequency $\omega_1$, $\omega_2$, $\omega_3$ and $\omega_4$. This case will also show how normalising by the number of points when using the WD makes a more robust metric for comparing data sets of different sizes.
    \item The second case study will remove $\omega_2$, and plot the manifold embedded in $\mathbb{R}^3$. Previous analysis of the Z24 bridge, has shown that a lot of the nonlinear features are present in the second natural frequency.  This case study is useful, as now that the natural frequency data are embedded in three dimensions, the data can be plotted and visualised, whilst showing that topological arguments are still valid in lower-dimensional shadows.
    \item The third case study shows the robustness of the manifolds to a linear dimension-reduction algorithm. In this case, the four-dimensional embedding of the data will be compressed down to two and three dimensions \cite{wold1987principal}. The reduced-dimension manifolds are then analysed with TDA.
\end{enumerate}

\subsubsection{Raw Natural Frequency Case}
For the first case, a full walk-through of the calculation procedure will be displayed. For the succeeding cases, this will be omitted to limit repetition. This case will take the first four natural frequencies obtained from the Z24 data set \cite{peeters2001one}. The natural frequencies will be represented by a point in four-dimensional space. It is believed that the introduction of damage will change the shape of the manifold in a more substantial way than the change in temperature. TDA can be used here to compare between the different shapes of the partitioned manifolds.

The persistent homology of the manifolds are calculated accordingly. The partitions are not all of the same size. The warm data set is the largest data set, and this will be randomly split in half to form two data sets, the original warm data set will remain in the analysis. The second can be used to verify the results; the two new random subsets will have a very similar topological structure, as they are all sampled from the same manifold. There will be slight differences because of topological noise formed in the persistence intervals from missing points in the smaller samples. These effects should be negligible when over the true global structure of the manifold, there are enough points to adequately describe the topology. For smaller partitions, the absence of many missing points will affect the topology in a significant way.

\begin{table}[H]
    \centering
    \begin{tabular}{lcccccc}
    \hline
    \textit{\textbf{}}        & \textit{\textbf{Freezing}} & \textit{\textbf{Cold}} & \textit{\textbf{Warm}} & \textit{\textbf{Damage}} & \textit{\textbf{Warm1}} & \textit{\textbf{Warm2}} \\ \hline
    \textit{\textbf{Freezing}} & 0.00 & 9.39 & 22.92 & 10.62 & 12.62 & 12.50 \\
    \textit{\textbf{Cold}}     & 9.39 & 0.00 & 21.47 & 5.35 & 8.78 & 8.50 \\
    \textit{\textbf{Warm}}     & 22.92 & 21.46 & 0.00 & 23.44 & 14.26 & 14.51 \\
    \textit{\textbf{Damage}}  & 10.62 & 5.35 & 23.44 & 0.00 & 10.09 & 9.59 \\
    \textit{\textbf{Warm1}}    & 12.62 & 8.78 & 14.26 & 10.09 & 0.00 & 1.80 \\ 
    \textit{\textbf{Warm2}}    & 12.50 & 8.50 & 14.51 & 9.59 & 1.80 & 0.00\\ \hline
    \end{tabular}
    \caption{WD values over each partition.}
    \label{tab:wass4d}
\end{table}
Table \ref{tab:wass4d} shows the WDs over the different partitions of the data set. These values are relatively uninteresting, as the WD values are a factor of the number of points present in the data. This effect can be seen clearly between the warm and the warm subsets, as all the WD values for the full warm data set are roughly twice the size of the two random subsets. This effect shows that the number of points present in the point cloud is linked to the size of the WD value.

As a more informative measure, the WD values can be summed along the rows. Summing along the rows, gives an understanding of how different a data set is from all the others. The larger the value, the more different it is. As well as this, one can normalise by the number of points present in the data set; this gives a normalised value, independent of size of the data set. This measure then acts as a metric to discriminate between manifold shapes for manifolds with a varying number of points present.

\begin{figure}[H]
    \centering
    \subfloat[{Raw.}]{\includegraphics[width = 0.45\textwidth]{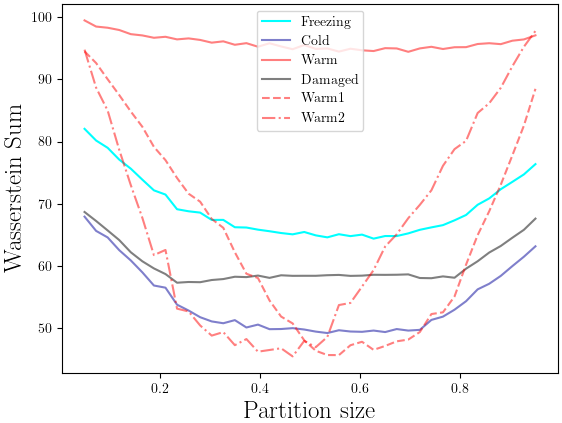}}
    \qquad
    \subfloat[{Scaled.}]{\includegraphics[width = 0.45\textwidth]{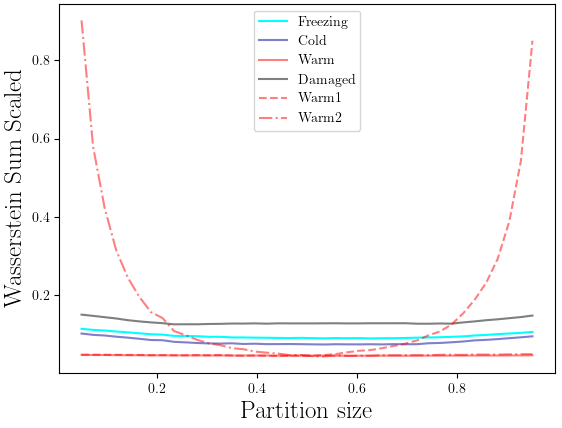}}
    \qquad
    \caption{Size of the WD values depending on the size of the warm partition size.}
    \label{fig: varyingpartition_size}
\end{figure}

Here, the change in the partition size and its effect on the WD value will be discussed, (Figure \ref{fig: varyingpartition_size}). The left-hand plot in Fig. \ref{fig: varyingpartition_size} shows how  the sum of the WD values for any data set varies as the partition size of the warm data set changes.  It can be seen that the very cold, cold, warm, and damage sets are all changing proportionally, as the partition size is varied, as is expected. On the other hand, the two warm subsets vary a lot as the partition size changes. For small partition sizes, the topology of the manifold is subsampled to an extreme and a likeness between the manifolds cannot be established.

The right-hand plot of Fig. \ref{fig: varyingpartition_size} shows the sum of all the WDs for a data set, and then scaled by the number of points presents in the data set. This figure shows how dividing by the number of points almost perfectly maps the warm subsets to the full warm set, in the regions where there are enough points to adequately describe the topology.

\begin{table}[H]
\centering
\begin{tabular}{lccc}
    \hline
    & \textit{\textbf{WD Sum}} & \textit{\textbf{Number of Points}} & \textit{\textbf{Scaled WD Sum}} \\ \hline
    \textit{\textbf{Freezing}}   & 68.050 & 720  & 0.095 \\ 
    \textit{\textbf{Cold}}       & 53.481 & 666  & 0.080 \\
    \textit{\textbf{Warm}}       & 96.584 & 2089 & 0.046 \\
    \textit{\textbf{Damage}}     & 59.095 & 457  & \textbf{0.129} \\
    \textit{\textbf{Warm1}}      & 47.549 & 1044 & 0.046 \\
    \textit{\textbf{Warm2}}      & 46.903 & 1045 & 0.045 \\ \hline
\end{tabular}
\caption{Summed and scaled WD data.}
\label{tab: summedndscaledwass}
\end{table}
As can be seen here, the damage case is clearly the most different in terms of manifold structure, compared to the other data sets. The freezing data are the next most distinct. 

For reference, if the partitions are not included in the analysis, results are obtained as in Tables \ref{tab: wassvalsnosubs} and \ref{tab: scalednosubs}. The damage manifold comes out as less different from the other values, as this now contains a greater weight in the analysis. When the warm subsets were included, this gave a much larger weight to this condition. Despite this, the damage manifold still comes out as the most topologically-dissimilar manifold.

\begin{table}[h]
\centering
\begin{tabular}{lcccc}
    \hline
& \textit{\textbf{Freezing}} & \textit{\textbf{Cold}} & \textit{\textbf{Warm}} & \textit{\textbf{Damage}} \\ \hline
    \textit{\textbf{Freezing}}  & 0.00  & 9.39 & 22.92 & 10.62 \\
    \textit{\textbf{Cold}}    & 9.39  & 0.00 & 21.47 & 5.35 \\
    \textit{\textbf{Warm}}    & 22.92 & 21.46  & 0.00 & 23.44 \\
    \textit{\textbf{Damage}} & 10.62 & 5.35 & 23.44 & 0.00 \\ \hline
\end{tabular}
\caption{WDs with subsets not included.}
\label{tab: wassvalsnosubs}
\end{table}

\begin{table}[h]
\centering
\begin{tabular}{lccc}
\hline
 & \textit{\textbf{WD Sum}} & \textit{\textbf{Number of Points}} & \textit{\textbf{Scaled WD Sum}} \\ \hline
\textit{\textbf{Freezing}}   & 42.922 & 720  & 0.060 \\ 
\textit{\textbf{Cold}}       & 36.203 & 666  & 0.054 \\ 
\textit{\textbf{Warm}}       & 67.817 & 2089 & 0.032 \\ 
\textit{\textbf{Damage}}     & 39.411 & 457  & \textbf{0.086}  \\ \hline
\end{tabular}
\caption{Scaled summed WDs with subsets not included.}
\label{tab: scalednosubs}
\end{table}

\subsubsection{3D Shadow: $\omega_1$, $\omega_3$, and $\omega_4$}
Previous analysis of the Z24 Bridge data has determined that $\omega_2$ contains the most nonlinear features. The analysis presented in the previous section can be repeated here, but after eliminating $\omega_2$ and plotting the manifold in three dimensions. This restriction acts as a visual aid as the data can now be plotted; it also presents a case where a lower-dimensional shadow is taken of the data, and the topology is preserved over this type of subsampling.

\begin{minipage}{0.48\textwidth}
\begin{figure}[H]
    \centering
    \includegraphics[width=0.9\textwidth]{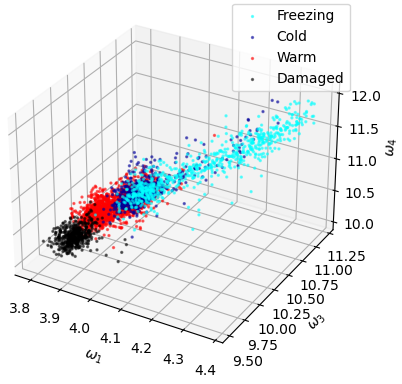}
    \caption{$\omega_1$, $\omega_3$, $\omega_4$ 3D plot, with the colours specifying data partitions.}
    \label{fig:omega134}
\end{figure}
\end{minipage}%
\hfill
\begin{minipage}{0.45\textwidth}
\begin{table}[H]
    \centering
    \begin{tabular}{lp{1cm}p{1.7cm}p{1.1cm}}
    \hline
    \textit{\textbf{}}        & \textit{\textbf{WD Sum}} & \textit{\textbf{Number of Points}} & \textit{\textbf{Scaled WD}} \\ \hline
    \textit{\textbf{Freezing}}   & 43.373                     & 720                                 & 0.060                         \\
    \textit{\textbf{Cold}}    & 32.461                     & 666                                 & 0.049                         \\ 
    \textit{\textbf{Warm}}    & 54.486                     & 2089                                & 0.026                         \\ 
    \textit{\textbf{Damage}} & 41.078                     & 457                                 & \textbf{0.090}                         \\ 
    \textit{\textbf{Warm1}}   & 29.190                     & 1044                                & 0.028                         \\ 
    \textit{\textbf{Warm2}}   & 29.013                     & 1045                                & 0.028                         \\ \hline
    \end{tabular}
    \centering
    \caption{WD values with the highly nonlinear $\omega_2$ removed.}
\end{table}
\end{minipage}%
\\

This example shows the resilience of the TDA procedure; how a lower-dimensional shadow still contains enough information to distinguish the damage data partition as the most different.

\subsubsection{Principal Component Analysis}

In this section, a linear dimension-reduction algorithm, called Principal Component Analysis (PCA) \cite{wold1987principal} will be applied to the data. Naturally, when reducing the dimension of the data, some information will be lost. The $4D$ space will be reduced to the principal components present in $3D$ and $2D$; the results of both will be presented. This demonstration will show how the data's topological structure is preserved over linear transformations, and how the accuracy degrades as the dimension reduction becomes more distant from the true embedding dimension.

The first principal component can be thought of as the direction that maximises the variance of the data into a projected space. The successive principal components are the ones that maximise the variance of the data into the projected space that are also orthonormal to all the previous components.

\begin{figure}[H]
    \centering
    \subfloat[{Regular torus.}]{\includegraphics[width = 0.3\textwidth]{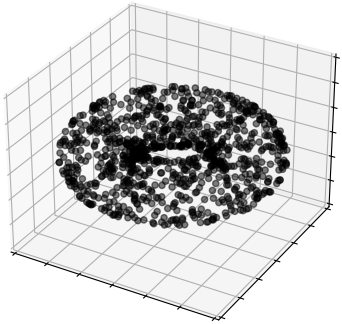}}
    \qquad
    \subfloat[{First two principal components.}]{\includegraphics[width = 0.3\textwidth]{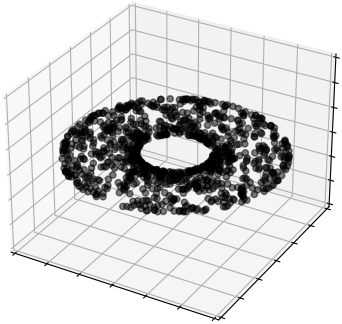}}
    \qquad
    \subfloat[{First principal component.}]{\includegraphics[width = 0.3\textwidth]{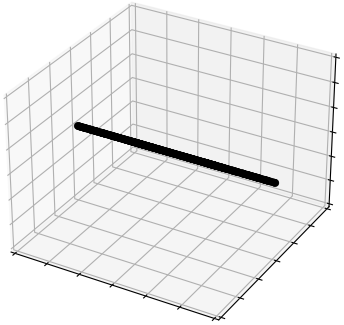}}
    \qquad
    \caption{Torus principal components.}
    \label{fig: torusPCA}
\end{figure}

To clarify how the topological structure is altered when taking the principal components, a case will be considered that uses some points randomly sampled from a torus. The example is shown in Fig. \ref{fig: torusPCA}. This figure shows that as the PCA embedding dimension is reduced, topological information is being lost. When going from a $3D$ embedding to a $2D$ embedding, the volume enclosed by the torus is lost. Following this progression, when going from $2D$ to $1D$, the embedding is simply a straight line with a distribution of points weighted at the ends. These changes in the embedding dimension mean topological information in the dimensions higher than the current embedding are lost, i.e. in Fig. \ref{fig: torusPCA} (b) the $3D$-volume is lost, and in Fig. \ref{fig: torusPCA} (c), the $2D$-hole is lost. Despite this loss of information, the topology captured in the remaining principal components is still adequately represented, in Fig. \ref{fig: torusPCA} (b), the $2D$-hole is still visible, and in (c), the manifold still consists of one connected component. 

The first PCA example presented here on the Z24 Bridge data set, is for the dimension reduction from $4D$ space into $3D$ space. As can be seen in Fig. \ref{fig:PCA43}, calculating the PCA of the data has still kept the clusters in somewhat separate regions of space. Under close inspection, the warm, cold, and freezing all seem to be single clusters. Whereas, the damage cluster appears to be comprised of two clusters positioned very close to one another. This property is one that will result in the topology of the damage partition being different to the other clusters.

\begin{minipage}{0.5\textwidth}
\begin{figure}[H]
    \centering
    \includegraphics[width=0.9\textwidth]{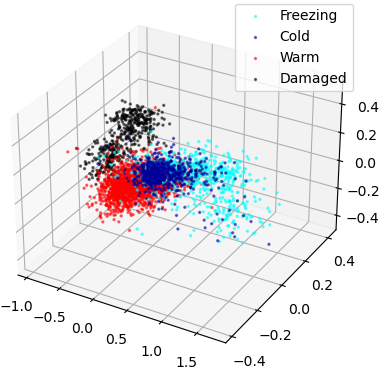}
    \caption{A visualisation of the first three principal components of Z24 natural frequency data.}
    \label{fig:PCA43}
\end{figure}
\end{minipage}%
\hfill
\begin{minipage}{0.45\textwidth}
\begin{table}[H]
    \centering
    \begin{tabular}{lp{1cm}p{1.7cm}p{1.1cm}}
    \hline
    \textit{\textbf{}}        & \textit{\textbf{WD Sum}} & \textit{\textbf{Number of Points}} & \textit{\textbf{Scaled WD}} \\ \hline
    \textit{\textbf{Freezing}}   & 56.763 & 720  & 0.079 \\
    \textit{\textbf{Cold}}       & 42.665 & 666  & 0.064 \\
    \textit{\textbf{Warm}}       & 69.743 & 2089 & 0.033 \\
    \textit{\textbf{Damage}}     & 44.878 & 457  & \textbf{0.098} \\
    \textit{\textbf{Warm1}}      & 36.561 & 1044 & 0.035 \\
    \textit{\textbf{Warm2}}      & 37.414 & 1045 & 0.036 \\ \hline
\end{tabular}
\caption{WD values for data reduced from $4D$ to $3D$.}
\label{tab:PCA43}
\end{table}
\end{minipage}%
\\

Performing the PCA of the data is interesting, as it shows that the topology and persistent homology is conserved over the first three principal components. As can be seen in Table \ref{tab:PCA43}, using the principal components of the data still results in the same orderings between the different partitions.

Now for a larger dimension reduction, into $2D$ space. This result still preserves the topology, although the WD values are a lot less distinct between the different cases. This result is because more topological information is being lost, as the data is compressed down to only the first two principal components. Despite the loss of information, the ordering from the scaled WD calculations still remains the same.

\begin{minipage}{0.5\textwidth}
\begin{figure}[H]
    \centering
    \includegraphics[width=0.9\textwidth]{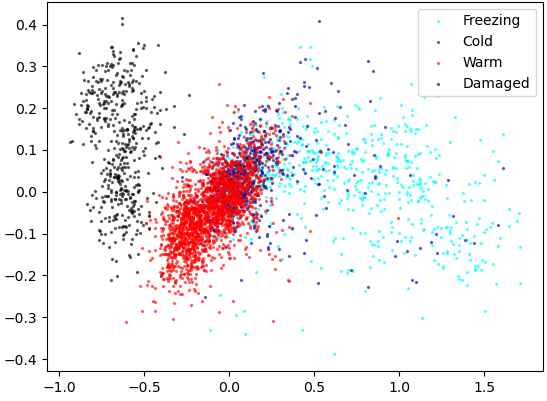}
    \caption{A visualisation of the first two principal components of Z24 natural frequency data.}
    \label{fig:PCA42}
\end{figure}
\end{minipage}%
\hfill
\begin{minipage}{0.45\textwidth}
\begin{table}[H]
\centering
\begin{tabular}{lp{1cm}p{1.7cm}p{1.1cm}}
    \hline
    \textit{\textbf{}}        & \textit{\textbf{WD Sum}} & \textit{\textbf{Number of Points}} & \textit{\textbf{Scaled WD}} \\ \hline
    \textit{\textbf{Freezing}}   & 28.074 & 720 & 0.039 \\ 
    \textit{\textbf{Cold}}       & 19.816 & 666 & 0.030 \\ 
    \textit{\textbf{Warm}}       & 25.156 & 2089 & 0.012 \\
    \textit{\textbf{Damage}}    & 19.310 & 457  & \textbf{0.042} \\
    \textit{\textbf{Warm1}}      & 16.637 & 1044 & 0.016 \\
    \textit{\textbf{Warm2}}      & 16.043 & 1045 & 0.015 \\ \hline
\end{tabular}
\caption{WD values for data reduced from $4D$ to $2D$.}
\label{tab:PCA42}
\end{table}
\end{minipage}%
\\

\subsection{Time-Delay Embedding}
In this section, the topology of the natural frequency manifold will be reconstructed from each of the $1D$ time series.
For the purposes of visualisation, the Z24 time series data will be embedded in $3D$, so that an intuitive understanding of the topology can be formed.

Figures \ref{fig:omega1TDE75} - \ref{fig:omega4TDE75} show the delay embeddings of the first four natural frequencies, projected into three dimensions. The embeddings for $\omega_1$, $\omega_3$ and $\omega_4$ show a similar structure at the delay value $\alpha=75$. Whereas, $\omega_2$ has a different topology defined around the colder temperatures. This can be inferred by the second mode having a more pronounced effect from the freezing temperature variations.

\begin{minipage}{0.45\textwidth}
\begin{figure}[H]
    \centering
    \includegraphics[width=0.9\textwidth]{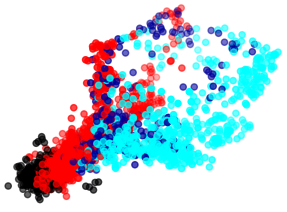}
    \caption{$\omega_1$ Time Delay Embedding $\alpha=75$.}
    \label{fig:omega1TDE75}
\end{figure}
\end{minipage}%
\hfill
\begin{minipage}{0.45\textwidth}
\begin{figure}[H]
    \centering
    \includegraphics[width=0.9\textwidth]{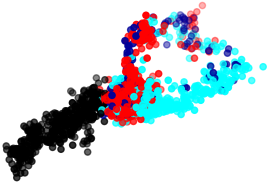}
    \caption{$\omega_2$ Time Delay Embedding $\alpha=75$.}
    \label{fig:omega2TDE75}
\end{figure}
\end{minipage}%
\\

\begin{minipage}{0.45\textwidth}
\begin{figure}[H]
    \centering
    \includegraphics[width=0.9\textwidth]{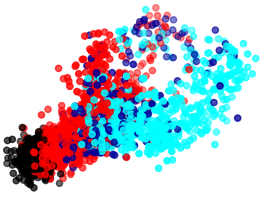}
    \caption{$\omega_3$ Time Delay Embedding $\alpha=75$.}
    \label{fig:omega3TDE75}
\end{figure}
\end{minipage}%
\hfill
\begin{minipage}{0.45\textwidth}
\begin{figure}[H]
    \centering
    \includegraphics[width=0.9\textwidth]{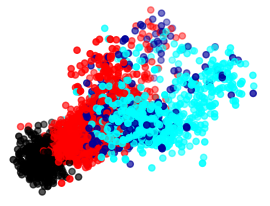}
    \caption{$\omega_4$ Time Delay Embedding $\alpha=75$.}
    \label{fig:omega4TDE75}
\end{figure}
\end{minipage}%
\\

When delaying the time series, this means that the delayed dimensions will have different temperature readings from the one-dimensional time series data used to form the embedding. The value for the temperature used to represent the point is the one from the initial time series data. The other dimensions will have their own respective temperatures. The classes used for the temperatures in these plots are a little less certain than in previous cases. This classification isn't too much of a problem, as the change in temperature is assumed to be a smooth change over each day, and as the value of $\alpha=75$ is relatively small compared to the size of the data set, the time delay can be inferred as a local change.

\section{Conclusion}
The focus of this paper was to present an overview of TDA to the structural dynamics community, with a variety of use cases that highlight its depth and borderline limitless capabilities. 

For the case of attractors, the calculation of the dimension of some common attractors was facilitated. A novel method for determining the optimal delay for constructing an attractor's topology from a $1D$ time series was also presented.

With respect to the Z24 data set, topological methods have been able to single out the damage data partition as the most topologically dissimilar. However, further analysis on topological methods for damage detection would need to be explored to understand the true limits and possibilities of TDA in SHM. An insight into the data structure provides powerful insight into the operating conditions of a machine or structure.

Future work on TDA, will look at different topological methods; aside from using the Wasserstein distance as a metric, other case studies will also be considered. A further journal paper, which extends on the ideas presented here will be submitted.

\section*{Acknowledgements}
The authors would like to thank the UK EPSRC for funding via the Established Career Fellowship EP/R003645/1 and the Programme Grant EP/R006768/1.

\newpage
\bibliographystyle{unsrt}
\bibliography{JP_7_21_TG}

\newpage
\listoffigures
\listoftables

\end{document}